# Barcode and QR Code Object Detection: An Experimental Study on YOLOv8 Models

Kushagra Pandya
*Dept.of Computer Engineering (AI)*
*Marwadi University*
Rajkot, India
kushagra.pandya110024@marwadiuniversity.ac.in

Heli Hathi
*Dept.of Computer Engineering*
*Marwadi University*
Rajkot, India
heli.hathi114884@marwadiuniversity.ac.in

Het Buch
*Dept.of Computer Engineering*
*Marwadi University*
Rajkot, India
het.buch114404@marwadiuniversity.ac.in

Ravikumar R N
*Dept. of Computer Engineering*
*Marwadi University*
Rajkot, India
ravikumar.natarajan@marwadieducation.edu.in

Shailendrasinh Chauhan
*Dept. of Computer Engineering*
*Marwadi University*
Rajkot, India
shailendrasinh.chauhan@marwadieducation.edu.in

Sushil Kumar Singh
*Dept. of Computer Engineering*
*Marwadi University*
Rajkot, India
sushilkumar.singh@marwadieducation.edu.in

*Abstract*— This research work dives into an in-depth evaluation of the YOLOv8 (You Only Look Once) algorithm's efficiency in object detection, specially focusing on Barcode and QR code recognition. Utilizing the real-time detection abilities of YOLOv8, we performed a study aimed at enhancing its talent in swiftly and correctly figuring out objects. Through large training and high-quality-tuning on Kaggle datasets tailored for Barcode and QR code detection, our goal became to optimize YOLOv8's overall performance throughout numerous situations and environments. The look encompasses the assessment of YOLOv8 throughout special version iterations: Nano, Small, and Medium, with a meticulous attention on precision, recall, and F1 assessment metrics. The consequences exhibit large improvements in object detection accuracy with every subsequent model refinement. Specifically, we achieved an accuracy of 88.95% for the nano model, 97.10% for the small model, and 94.10% for the medium version, showcasing the incremental improvements finished via model scaling. Our findings highlight the big strides made through YOLOv8 in pushing the limits of computer vision, ensuring its function as a milestone within the subject of object detection. This study sheds light on how model scaling affects object recognition, increasing the concept of deep learning-based computer creative and prescient techniques.

Keywords— Object Detection, YOLOv8, Deep Learning, Barcode Detection, QR Code Detection

## I. INTRODUCTION

A barcode is similar to "optical Morse code" for unique object identification, with black lines and white spaces printed on labels [2]. Barcodes, both 1D and 2D, such as EAN (European Article Number) and UPC (Universal Product Code) for retail items and QR codes for linking print to digital media, are being used in a variety of scenarios, ranging from local stores to marketing campaigns [3]. Real-world situations involve localization, image preprocessing, and decoding steps, all crucial for retrieving encoded information [4]. Reading barcodes from cell phones often requires carefully framed pixels, limiting their usefulness [5]. Due to high demand, fast and accurate barcode detection is crucial [1]. So, by using Deep-Learning based object detection algorithms like YOLOv8, offers promising answers. The state-of-the-art one-stage object detection network YOLOv8 is chosen as the benchmark in this paper. YOLOv8s combines the advantages of fast detection speed and a small number of network parameters.

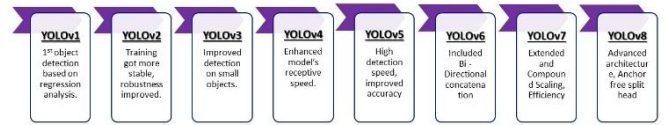

Fig. 1 Versions of YOLO

### A. Advantages of YOLOv8

- **Better mAP:** YOLOv8 surpasses previous versions in mean average precision, excelling in detecting smaller objects and challenging conditions.
- **Efficiency:** Achieves high performance with less parameters, potentially reducing learning occasions.
- **Developer-friendly:** Enhancements like multi-GPU aid, greater serialization simplifying improvement techniques.
- **Robust:** Handles scale variations and other challenges successfully, ensuring dependable outcomes across diverse scenarios.
- **Faster Inference:** YOLOv8 gives you fast inference speeds, perfect for real-time packages which include self-driving motors and video evaluation.

### B. Applications of YOLOv8

- **Autonomous Vehicles:** Detect objects on roads in real time, crucial for independent automobile systems.
- **Security and Surveillance:** Monitors high-protection places like airports, enhancing surveillance structures.
- **Medical Imaging:** Aids in decoding medical scans for prognosis and remedy planning.
- **Industrial Automation:** Tracks gadgets on manufacturing strains, detects defects, ensures best management.

### C. Goals of the Research

- Evaluate the overall performance of YOLOv8 in object detection, especially in appropriately across smaller gadgets and in hard conditions.
- Examine the performance of YOLOv8 in parameter utilization and training times, aiming to streamline the model and improve computational efficiency.
- Analyse the flexibility offered by way of special variants of YOLOv8 (YOLOv8n, YOLOv8s, YOLOv8m) for various needs, balancing elements

along with accuracy, velocity, and useful resource utilization.

- Provide insights into the strengths and boundaries of YOLOv8 based on empirical evaluations and actual international use instances, aiming to manual researchers and practitioners in leveraging this technology effectively.
- Contribute to the continuing discourse surrounding similar approaches in Computer Vision, advancing the knowledge of modern-day object-detection techniques and their effects on numerous programs.

## II. LITERATURE REVIEW

Numerous studies and contributions have made important advances in the field of object detection. Notably, the YOLO (You Only Look Once) strategy stands out as the latest method, valued for its fast computational speed and low parameter needs. A thorough review of recent research, as shown in Table 1, demonstrates a strong dependence on the YOLO framework for object detection tasks. Some studies improve their models by incorporating specific modules (e.g., [15] and [17]) or combining them with complementary algorithms (e.g., [13], [14], [19], [20]). While numerous datasets serve as the foundation for object detection tasks, there is a clear trend toward the use of custom datasets customized to specific research aims (e.g., [11], [13], [16], [20], [22]). Furthermore, with rare exceptions, most studies use a learning rate of 0.01 for their investigations ([10], [16], [20]). Further, a common choice of epochs falls between 200 and 500, indicating efforts to improve model efficiency. The most common evaluation measures used in these investigations are accuracy and mAP (mean average precision). Some studies, such as Ref No. 18 (1080×1080) and Ref No. 22 (416×416), use a different image size than the standard 640×640. These findings demonstrate the importance of YOLO-based techniques in driving progress in object detection research. Abbreviations Used: LR – Learning Rate, BS – Batch Size, Opt. – Optimizer.

## III. ARCHITECTURE AND METHODOLOGY

### A. Architecture

The YOLOv8 architecture includes 5 main steps such as input, down sampling, fusion, detection, and output. Further, the steps are briefed in the Fig. 2.

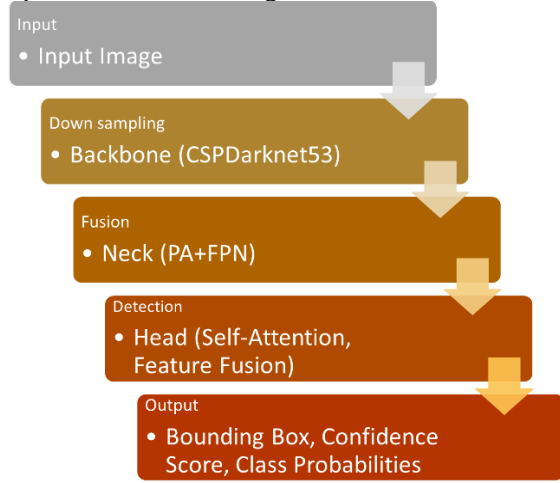

Fig. 2 YOLOv8 Architecture

TABLE I. LITERATURE REVIEW AND RELATED STUDIES

| Ref No. | Dataset Used | Model | Parameters Used | Accuracy |
|---|---|---|---|---|
| [10] | FMD Dataset | YOLOv8s | LR=0.1, BS=16, Epochs=200 | mAP(0.95) - 0.93 |
| [11] | US Dataset along with custom Dataset | YOLOv8n, YOLOv8m | LR=0.01, Epochs=50, Opt.=SGD | 99.8% |
| [12] | TT100k Dataset | YOLOv8 (Customized) | LR=0.01, BS=32, Epochs=200 | 80.8% |
| [13] | Custom Dataset | YOLOv8 (n, m, s) and CenterNet model | LR=0.01, BS=2, Opt.=SGD | 95% |
| [14] | Pascal VOC Dataset | SSB-YOLO | LR = 0.01, BS=16, Opt.=SGD, Epochs=300 | 87.5% |
| [15] | COD10K, CAMO Dataset | YOLOv8+EFM+ EEM | LR=0.01, BS=32, Opt.=SGD, Epochs=500 | 70.5% 64% |
| [16] | Weeds Public Dataset along with Custom Dataset | EfficientDet, YOLOv5m, YOLOv6l, YOLOv7, YOLOv8l | LR=0.0001, BS=8, Epochs=100, Opt.=SGD | YOLOv7– 89% (Weeds) YOLOv7 – 54% (Custom) YOLOv8 – 90% (Weeds) YOLOv8 – 53.9% (Custom) |
| [17] | BDD100K, NEXET Dataset | YOLOv8 (Modified: SnakeVision) | LR=0.01, BS=32, Epochs=1000, Layers=168 | mAP(50-95)-0.449 mAP(50-95)-0.495 |
| [18] | Obj. Detection and localization of image Dataset | YOLOv8 (with Transfer Learning) | LR=0.01, Epochs=20 | mAP(50) of 0.874 at 66.7 FPS |
| [19] | LSI X-Ray, OPI X-Ray Dataset | SC-YOLOv8 | Not Provided | LSI X-Ray-82.7% OPI X-Ray-89.2% |
| [20] | Custom Dataset | SHFP-YOLO | LR=0.001, BS=16, Epochs=500, Opt.=SGD | mAP(50-95)-0.42 |
| [21] | MASK,Face (FDDB) Dataset | YOLOv8 | Epochs = 100 | 84.6% |
| [22] | Custom Dataset | YOLOv8 | Epochs=100, BS=64 | 84.7% |
| [23] | Open Source Dataset | YOLOv8n | Not Provided | >=95% |
| [24] | RailSem19 Dataset | YOLOv8x | LR=0.01, Opt.=SGD | mAP(50)-0.88 |

Following YOLOv5 in 2020, Ultralytics released YOLOv8 in January 2023. Labelling, training, and deployment are integrated with CLI and PIP packages. YOLOv8 employs online image augmentation and mosaic augmentation, which mixes four photos every iteration, to increase accuracy on training. These techniques improve model training. Larger feature maps and more efficient convolutional networks make YOLOv8 better [6]. A backbone (feature extraction), neck (multi-level feature combining), and head (classification and localization predictions) make up YOLOv8 [7]. Ignoring the top-down up-sampling phase of the PAN-FPN in the neck layer, YOLOv8 replaces the C3 module with the gradient-enriched C2f [8].

Modules Conv, C2f, and SPPF form the system's backbone. Modules collaborate to achieve tasks. Essential qualities are determined by two Conv modules. C2f module improves these traits. SPPf refines features by evaluating various picture sections. Semantic comprehension and deep analysis are separated in the PAFPN backbone structure. After that, the primary module creates three separate maps of various picture sections. Head module decouples classification and localization and substitutes anchor-based with anchor-free, decreasing model calculations and boosting convergence speed and efficacy [8]. This technique used a single neural network to identify objects. Building and shaping the algorithm on a complete picture is simple.

## IV. METHODOLOGY

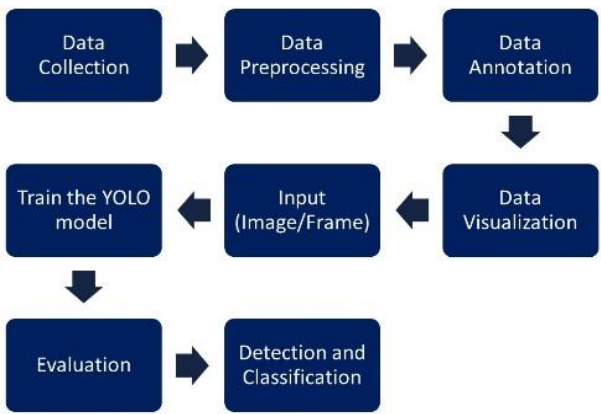

Fig. 3 Workflow Methodology

### A. Data Collection

The first part of the procedure involves data gathering, in which a custom dataset [9] of 31,078 images is selected from Kaggle and divided into training (28,696) and validation (2,382) images.

### B. Data Preprocessing

The obtained data is then pre-processed to clean and optimize it for training. Following pre-processing, data annotation is performed using RoboFlow.

### C. Data Annotations

We also performed various augmentations which included several augmentations such as rotation (-15° to 15°), blur (up to 3.25px), crop (minimum zoom - 0%, maximum zoom - 22%), and flip (both horizontal and vertical), among others.

### D. Data Visualization

The Fig 4. Shows the samples of custom-made Barcode and QR code samples.

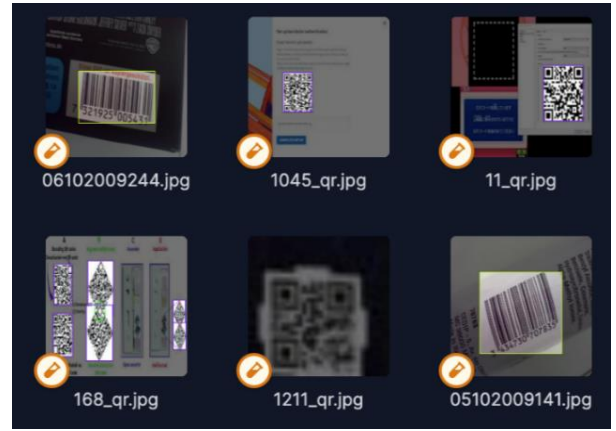

Fig. 4 Barcode and QR code dataset

### E. Training

After data pre-processing and annotation, the YOLOv8 model is used for training. Three variants of the YOLOv8 model are used for this purpose: YOLOv8n, YOLOv8s, and YOLOv8m, with a learning rate of 0.01 and training epochs set to 10. Image size taken for this study is 416×416.

TABLE II. TECHNICAL SPECIFICATION

| IDE | Jupyter Notebook |
|---|---|
| Environment | Python 3.10.13 |
| OS | Linux 5.15.133+ |
| CPU | Intel(R) Xeon(R) 2.00 GHz |
| CPU RAM | 31.36 GB |
| GPU RAM | 16 GB |
| GPU | Tesla P100-PCIE-16GB |
| *Training Time Consumed* | |
| Nano Model | 1 Hour 7 Minutes |
| Small Model | 1 Hour 37 Minutes 31 Seconds |
| Medium Model | 3 Hours 14 Minutes 8 Seconds |

### F. Evaluation

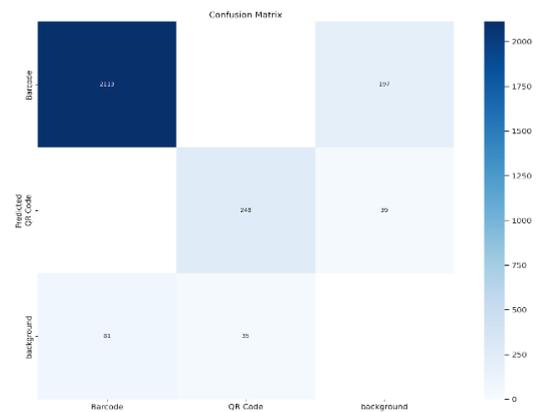

Fig. 5 Confusion Matrix of YOLOv8n

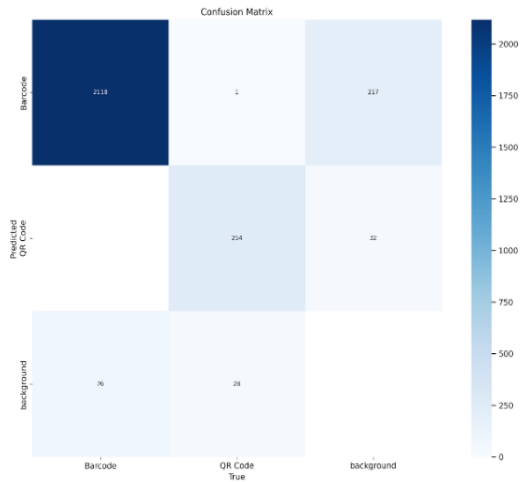

Fig. 6 Confusion Matrix of YOLOv8s

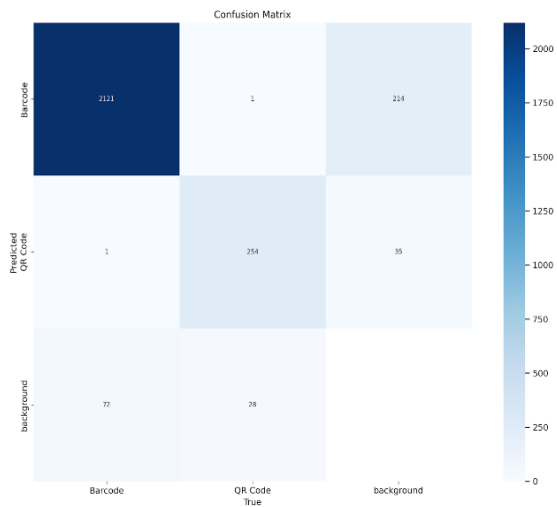

Fig. 7 Confusion Matrix of YOLOv8m

## V. RESULT ANALYSIS

As given in following table (Table 2) the performance of the three YOLOv8 variants (YOLOv8n, YOLOv8s, and YOLOv8m) in object detection tasks was evaluated based on several key metrics. YOLOv8m achieved the highest recall (0.9008) and accuracy (0.9410) among the models, indicating its effectiveness in correctly identifying objects. YOLOv8s followed closely with a recall of 0.8811 and an impressive accuracy of 0.9710. YOLOv8n exhibited slightly lower recall (0.8885) and accuracy (0.8895) compared to the other two models. YOLOv8s demonstrated the highest precision (0.8541), followed by YOLOv8m (0.8556) and YOLOv8n (0.8477).

However, when considering the balance between precision and recall, YOLOv8m achieved the highest F1 score (0.8785), indicating its ability to maintain a balance between precision and recall. YOLOv8s and YOLOv8n had F1 scores of 0.8663 and 0.8677, respectively. YOLOv8s achieved the highest mAP score (0.8967), followed closely by YOLOv8m (0.8901) and YOLOv8n (0.8788). The mAP50 and mAP50-95 scores were also highest for YOLOv8s, indicating its effectiveness in detecting objects across different thresholds. YOLOv8m exhibited the lowest training loss (0.9955) and validation loss (1.271) among the models, suggesting its robustness and ability to generalize well to unseen data. YOLOv8s had slightly higher losses (training loss: 1.268, validation loss: 0.9974), while YOLOv8n demonstrated the highest losses (training loss: 1.01, validation loss: 1.272).

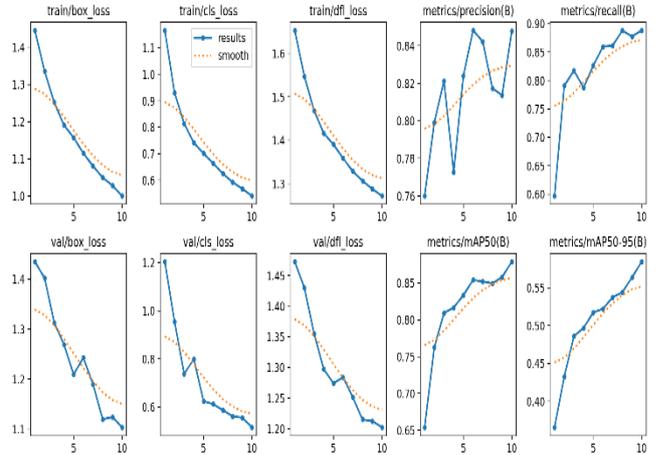

Fig. 8 Metrics Evaluation Graphs of YOLOv8n

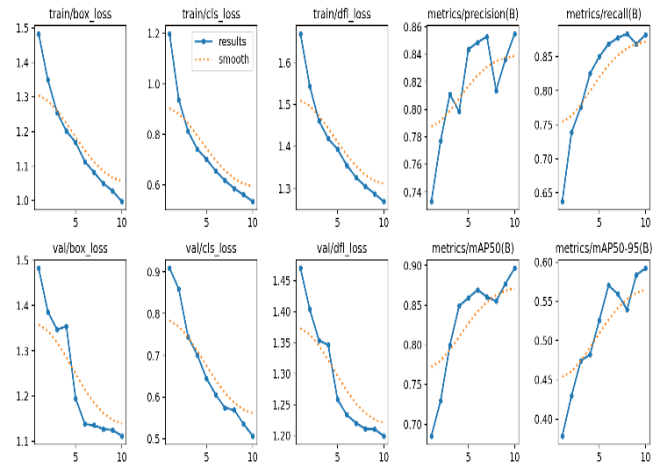

Fig. 9 Metrics Evaluation Graphs of YOLOv8s

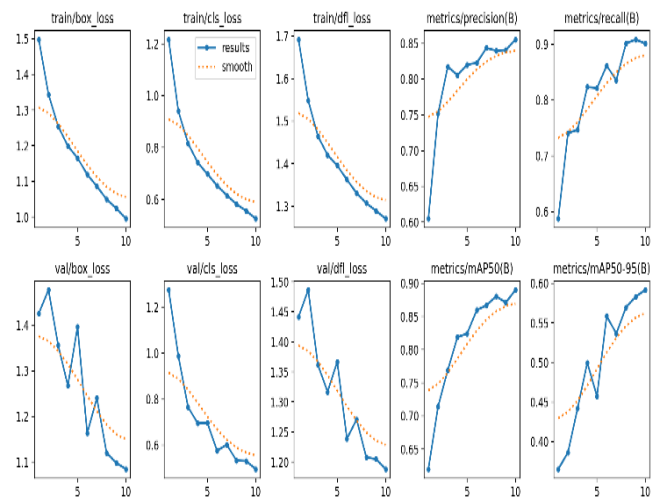

Fig. 10 Metrics Evaluation Graphs of YOLOv8m

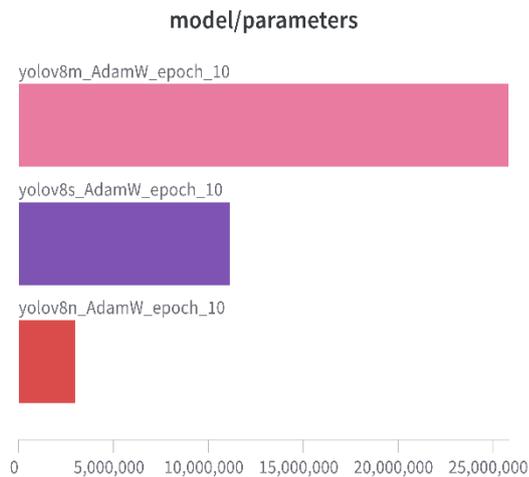

Fig. 11 Total Number of Parameters

Fig. 11 shows the number of parameters of each model during its training. From figure, the number of parameters grows notably as the size of the model increases. For YOLOv8n model, total parameters used are 3011238 parameters least between all models. For YOLOv8s the parameters increased to 11136374 parameters and for YOLOv8m the parameters were 25857378 parameters, largest of all.

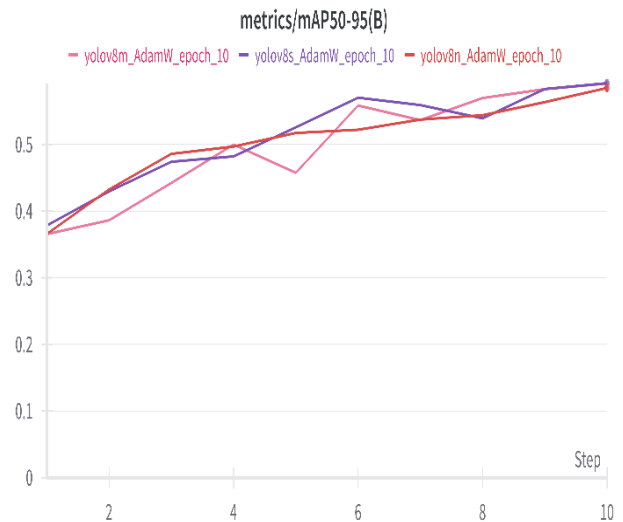

Fig. 13 mAP50-95 Metrics

Fig. 13 shows the mAP (Mean Average Precision) score with a threshold value of between 50% and 95% for more precise and accurate results. There are no major differences in mAP50-95 score for all the three models. (0.59 for all three YOLOv8 models). As all three models have a similar mAP50-95 score, it means they have similar overall accuracy in detecting and localizing objects at various IoU thresholds.

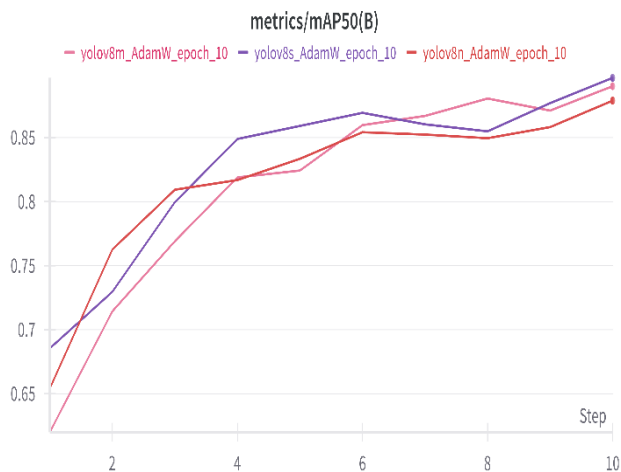

Fig. 12 mAP50 Metrics

Fig. 12 shows the mAP (Mean Average Precision) score with a threshold value of 50%. The mAP50 metric is commonly used to evaluate object detection algorithms. Observation reveals that all three YOLOv8 models have about equivalent mAP50 values, with YOLOv8s slightly outperforming the other two YOLOv8 models (YOLOv8n - 0.88 and YOLOv8m - 0.89 from Table III).

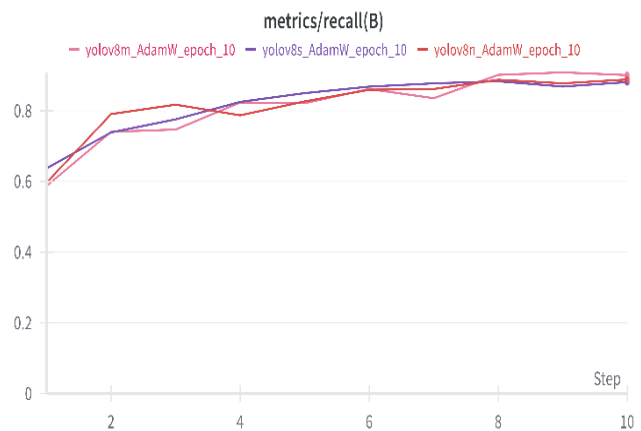

Fig. 14 Recall Metrics

Fig. 14 provides insights on the recall metrics of the YOLOv8 models. The graph shows occasional variations and intersections, demonstrating that the measures' relative performance might vary throughout epochs. However, the general trend indicates that the "yolov8m" slightly betters the other two in terms of recall for this model training.

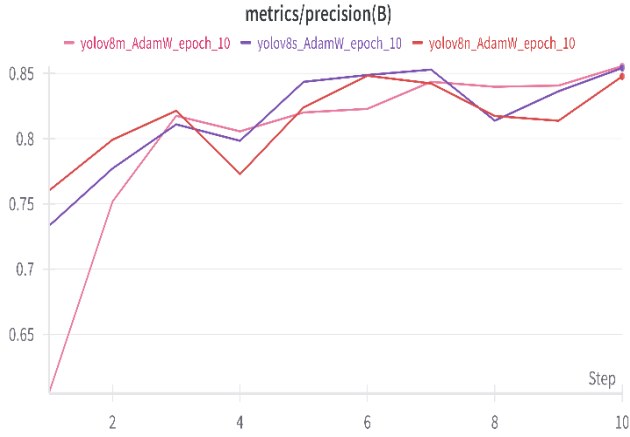

Fig. 15 Precision Metrics

Fig. 15 shows the precision performance of the three YOLOv8 models. Although there are significant variations at initial stage of training, (YOLOv8m having significant increment in precision) after further training all three variations of YOLOv8 has very similar precision at the end of training (Results provided in Table III).

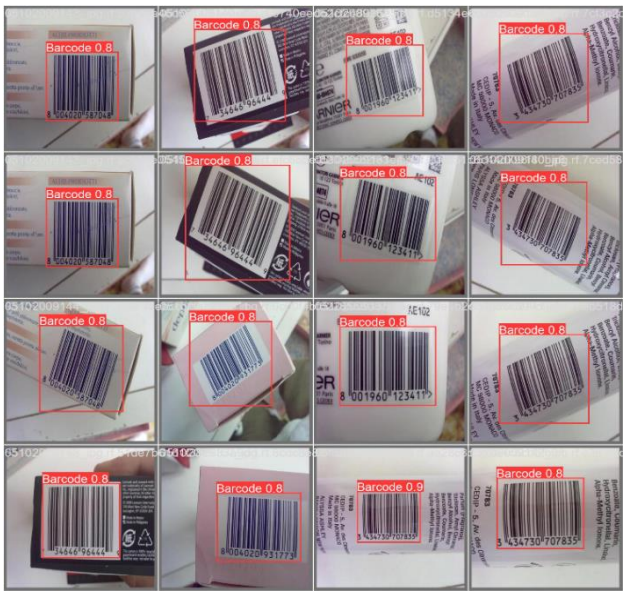

Fig. 16 Predicted Object Detection

Finally, after the model training, we feed the model with the images that model has not seen during its training. Based on model's prediction and confidence scores, we check which model is best suited for application in real world.

models which clearly outperforms the previous model Yolov7's performance.

## VI. CONCLUSION

Finally, this work evaluates YOLOv8's barcode and QR code recognition in images using a dataset that contains both categories. Experimentation shows that sophisticated YOLOv8 family designs improve detection. After hyperparameter fine-tuning, YOLOv8s had the maximum accuracy of 97.1%, making it best for this detection. Despite having the highest mAP, YOLOv8n may be overfitting. Despite having lesser accuracy than YOLOv8s, YOLOv8m has better generalization, as seen by its lowest losses. Creating a real-time online barcode and QR code identification software and improving model efficiency through algorithmic and hardware upgrades are future goals. These efforts aim to improve YOLOv8's application detection.

## ACKNOWLEDGMENT

We are thankful to Prof. Ravikumar R N, Prof. SS, and Dr. SKS for their guidance and insights throughout our project and paper. Their support was crucial in completing this work, which is a result of the Research Activities Club at the Department of Computer Engineering, Marwadi University, Rajkot, India.

TABLE III.    RESULT ANALYSIS TABLE

| Models | Recall | Accuracy | Precision | F1 Score | mAP | | Train Loss | | | Validation Loss | | |
|---|---|---|---|---|---|---|---|---|---|---|---|---|
| | | | | | mAP50 | mAP 50-95 | Classification | DFL | Box Loss | Classification | DFL | Box Loss |
| YOLOv8n | 0.89 | **0.89** | 0.85 | 0.87 | 0.88 | 0.59 | 0.54 | 1.01 | 1.27 | 0.52 | 1.20 | 1.10 |
| YOLOv8s | 0.88 | **0.97** | 0.85 | 0.87 | 0.90 | 0.59 | 0.53 | 1.27 | 1.00 | 0.51 | 1.20 | 1.11 |
| YOLOv8m | 0.90 | **0.94** | 0.86 | 0.88 | 0.89 | 0.59 | 0.53 | 1.00 | 1.27 | 0.50 | 1.19 | 1.09 |

The result table III shows the performance of Yolov8 with the accuracy 89% for nano, 97% for small and 94 for medium